\documentclass[runningheads]{llncs}
\usepackage[T1]{fontenc}
\usepackage[utf8]{inputenc}
\usepackage{amsmath}
\usepackage{caption,subcaption}
\usepackage{pgffor}
\usepackage{csquotes}

\usepackage{multirow}
\usepackage{microtype}
\usepackage[english]{babel}

\usepackage[table]{xcolor}
\usepackage{hyperref} 

\usepackage{booktabs}
\usepackage{tabularx}
\usepackage{ragged2e}
\usepackage{longtable} 
\usepackage{afterpage}
\usepackage{rotating}
\usepackage{enumitem}
\usepackage{graphicx}
\usepackage[most]{tcolorbox}
\usepackage{calc}
\usepackage{nicefrac}

\usepackage[export]{adjustbox}

\begin{document}

\title{Leveraging Morphology for Historical \\ Script Metrological Analysis}

%

\newif\ifanonym
\anonymfalse

\ifanonym
  \author{Anonymous Author(s)}
  \authorrunning{Anonymous Author(s)}
  \institute{} 
\else
  \author{
  Malamatenia Vlachou-Efstathiou\textbf{*}\inst{1,2}\orcidID{0000-0002-9397-356X} \and
  Raphael Baena\textbf{*}\inst{1}\orcidID{0000-0003-3214-3252} \and
  Dominique Stutzmann \inst{2}\orcidID{0000-0003-3705-5825}
  \and
  Mathieu Aubry\inst{1}\orcidID{0000-0002-3804-0193}}
  \authorrunning{}
  \institute{
  LIGM, École des Ponts et Chaussées, IP Paris, CNRS, France \and
  Institut de Recherche et d'Histoire des Textes, Paris, Île-de-France, France\\
  \email{\{first.last\}@enpc.fr}}
\fi


\maketitle 


\vspace{-2em}
\begin{figure}[h!]
\centering
\includegraphics[width=\textwidth]{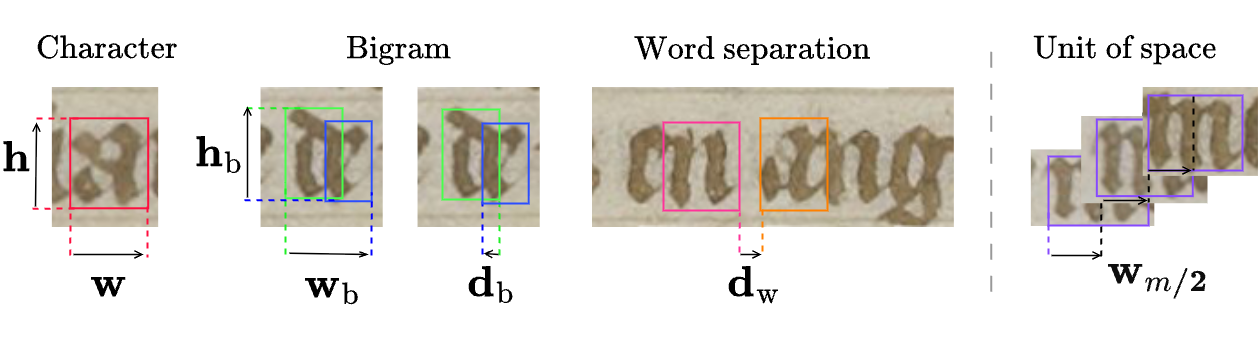}
\caption{Using character bounding boxes defined by the best fit between a prototypical character and a specific character instance, without human annotations, we enable stable and relevant paleographical measurements for characters, bigrams, and word separation. Our approach can be seen as bridging morphological and metrological paleographic analysis and scales to hundreds of units of analysis.}
\label{fig:metrics}
\vspace{-2em}
\end{figure}

\begin{abstract}
Advances in handwritten text recognition have enabled large-scale transcription of historical documents, but still provide limited access to interpretable visual measurements for paleography, the study of historical scripts. In this paper, our main insight is that morphological script analysis, in particular the capacity to learn character prototypes from line-level transcriptions, enables the definition of scalable, meaningful, and stable paleographic measurements. More precisely, we leverage a transformer-based detection architecture together with a prototype-based line reconstruction module to learn prototypical characters and their occurrence, deformation, and positioning.

Our contributions are twofold. First, we introduce a deep architecture and learning methodology that enables efficient character modeling with only line-level transcription supervision, significantly improving over the Learnable Typewriter baseline and enabling accurate character bounding box prediction, unlocking its potential for paleographic measurements. Second, we introduce and demonstrate the paleographical relevance of automatic measurements enabled by our architecture for characters, bi-grams, and spaces between graphical units. For this demonstration, we extend the annotations of the codex Paris, BnF, fr. 2813, commissioned in the late fourteenth century by Charles V and copied by four hands, to 160 pages. We visualize our measurements over these pages, showing how they enable us not only to differentiate graphical profiles, but also to discover and analyze subtle variations. This case study outlines the scalability of our approach and its frugality in terms of required training data, since a single column of text is sufficient to compute our measurements on each of the 160 pages. \\

Data and code for the method and the analysis are publicly available at: \url{https://malamatenia.github.io/morphology4metrology-analysis}.

{}
\end{abstract}
%
%
%
\section{Introduction}

Despite advances, paleographical analysis —understood as the study of the graphical properties of writing systems— remains largely dependent on expert paleographers' ability to memorize and compare forms. Although rigorous, this process is inherently subjective, which limits communicability, reproducibility, and scalability. In this paper, we introduce the technical and paleographic foundations of automatic measurements and demonstrate their paleographical relevance through a specific case study. Our main insight is to build a system that bridges two traditionally independent approaches, morphological and metrological analysis, and to use morphology to define consistent measurements.

For more than five decades, attempts have been made to translate paleographic expertise and the \textit{modus operandi} of paleographers into formalized metrological procedures~\cite{poulle1974paleographie,gilissen1975rapportmodulaire}. However, both manual and digital workflows still struggle to scale the extraction of standardized, trustworthy, measurable features of handwriting. 
In contrast, both semi-automatic methods~\cite{aiolli1999spi,ciula2005digital,mamatsis2023novel}and recent advances in generative modeling, such as the Learnable Typewriter~\cite{ltw2023} and its application to paleography, the Learnable Handwriter~\cite{lhw2024}, have demonstrated the relevance of extracting character prototypes directly from text lines for paleographical morphological analysis. To bridge this gap between metrological and morphological approaches, we designed paleographically relevant measures based on morphological analysis. Namely, we define boundaries of characters using the optimal alignment between character prototypes and specific character instances.

We build on the idea of the Learnable Handwriter~\cite{lhw2024} to jointly learn character prototypes and placements by reconstructing a dataset of text lines with their transcriptions, and to specialize prototypes for specific units of analysis. However, the Learnable Handwriter~\cite{lhw2024} suffers from limitations that prevent it from being used to obtain metrological information. Most importantly, it does not account for changes in character aspect ratio, which is critical for paleographic measurements, and its prototypes learn background and context, requiring early stopping of the training at a high character error rate (CER).

We overcome these limitations through an architecture and training methodology that bring together ideas from detection-based text recognition and prototype-based analysis-by-synthesis. We extend the Detection Transformer for Line Recognition (DTLR)~\cite{DTLR} approach with a reconstruction module, which utilizes bounding boxes to place and deform character prototypes on a predicted low-resolution background. We also adapt the prediction and reconstruction module to model accents and abbreviations and improve the memory efficiency of the reconstruction module. These changes lead to improved prototypes, with less noise and better use of the space of their canvas, compared to the Learnable Handwriter~\cite{lhw2024} baseline. More importantly, they enable us to introduce and compute paleographically relevant measures, visualized in Figure~\ref{fig:metrics}, on the proportionality of characters and the horizontal compression of the graphic chain, two critical properties of script. 

To demonstrate the effectiveness of our approach and measures, we develop a case study on the \textit{Grandes Chroniques de France} (Paris, BnF, fr. 2813), copied in the late fourteenth century by four hands, and for which we extend existing annotations~\cite{vlachou_efstathiou_2025grandes-chroniques-fr-2813_dataset} to 160 folios. We show that our method improves over the Learnable Handwriter both in terms of prototype quality and training stability. More importantly, we showcase the benefits of our measurements through visualizations, which not only allow us to identify the different hands but also highlight subtle inter- and intra-hand variations.

\vspace{-0.5em}
\subsubsection{Contributions.} To summarize, our contributions are as follows:
\begin{itemize}[leftmargin=*,topsep=2pt]
\item we bridge morphological and metrological paleographic analysis by defining measurements based on character prototype alignments, opening the way for more consistent and scalable metrological analysis;
\item we introduce an architecture and training strategy that brings together detection-based character recognition and prototype-based modeling, which improves results, stability, efficiency, and enables us to perform our paleographic measurements;
\item we demonstrate the effectiveness of our approach and measures through a case study, outlining their potential for fine-grained analysis and discovery beyond hand recognition.

\end{itemize}
\section{Related Work}

\subsection{Text recognition and modeling}
Our approach combines detection-based text recognition with prototype-based image reconstruction to obtain both interpretable prototypes and accurate character positioning. We first review text recognition paradigms, then reconstruction-based character modeling approaches, our contribution being at the intersection of these two lines of work.

\subsubsection{Text Recognition.}
In recent years, text recognition has been predominantly addressed using deep learning methods based on sequence-to-sequence formulations~\cite{diaz2021rethinking,htrvit,trocr}. Early approaches combine the Connectionist Temporal Classification (CTC) loss~\cite{CTC} with recurrent architectures such as CRNN~\cite{CRNN}, LSTMs~\cite{graves2005framewise,bluche2017gated}, or multidimensional LSTMs~\cite{voigtlaender2016handwriting,puigcerver2017multidimensional}. More recent attention-based and Transformer models significantly improve recognition performance~\cite{michael2019evaluating,kang2022pay,diaz2021rethinking,trocr,htrvit}. 

In contrast to sequence-to-sequence models limited to transcriptions, detection-based text recognition explicitly predicts character instances and their spatial locations, providing character-level annotations and geometric information -- position, width, height, aspect ratio -- that are essential for fine-grained paleographical analysis. Although no longer prevalent for Latin alphabets, this strategy remains important for other languages, particularly Chinese HTR~\cite{chineseexplicitsegm1,peng2019fast,yu2024approach}. Recently, DTLR~\cite{DTLR} adopted this approach by adapting DINO-DETR~\cite{DINO-DETR} to predict character bounding boxes and class labels. The authors show that, after pretraining on synthetic data, the model can be finetuned on real manuscripts using only a modified CTC loss, without any bounding box annotations. Remarkably, this finetuning yields meaningful bounding boxes despite the lack of spatial supervision. However, nothing ensures the consistency of these bounding boxes with respect to the characters, making it impossible to use them directly for paleographic measurements.

\subsubsection{Character Modeling and Reconstruction.}
Modeling characters has been explored to refine OCR predictions and obtain interpretable prototypes. Early systems such as Ocular~\cite{berg2013unsupervised} and related approaches~\cite{xu1999prototype,aradillas2021boosting} jointly model character shapes, ink rendering effects, and language priors. Other works~\cite{sinha2019unsupervised,jenckel2018transcription} fine-tune pretrained OCR models by optimizing a reconstruction loss using LSTM-based decoders or GANs. However, these methods often work only on datasets similar to their original training data, limiting their applicability to historical documents, and their components are rarely trained end-to-end~\cite{berg2013unsupervised,aradillas2021boosting}. None of these approaches has been leveraged for paleographical analysis.

More recently, building on sprite-based decomposition~\cite{monnier2021unsupervised,smirnov2021marionette}, the Learnable Typewriter~\cite{ltw2023} proposed an end-to-end architecture that jointly performs text recognition and text-line reconstruction by learning a prototype for each character along with its spatial transformations and class predictions. While this approach was leveraged by the Learnable Handwriter~\cite{lhw2024} for paleographical analysis~\cite{vlachou2025grandes-chroniques-fr-2813}, it simply pastes characters without enabling deformations, and thus gives no direct way to measure distances, width, height, or aspect ratio. 

In this work, we extend DTLR~\cite{DTLR} with a prototype-based reconstruction module, obtaining both interpretable character prototypes and measurements required for metrological analysis in a single unified framework. This approach is also significantly faster and more accurate than~\cite{lhw2024}.

\subsection{Computer-assisted approaches in paleography}

Paleographic research aims to describe, compare, and analyze manuscripts for tasks from the identification of scribal hands to the study of script types. 
Traditional paleographic methods, based on textual and graphical features, have been criticized for their subjectivity and the difficulty of reproducing and evaluating results~\cite{derolez2003palaeography}. This has motivated the development of quantitative and computational approaches, aiming at objective measurements and
reproducibility. 
We focus on computational approaches relying on graphical features and separate works that deal with character shapes ({morphology}) and measurements ({metrology}).
 

\subsubsection{Morphology.}

Most computational approaches focus on the shapes of letters and the distribution of allographs~\cite{gurrado_mesure_2013,stutzmann_conjuguer_2014}. 
Approaches based on image analysis aggregate multiple instances of a character to generate abstract prototypical forms representative of a hand or script type~\cite{muzerelle_analyse_2011,popovic2021artificial}, as in traditional {idealized} alphabets.
For example, the System of Palaeographical Inspection (SPI)~\cite{aiolli1999spi,ciula2005digital} generates average prototypes from semi-automatically segmented characters
, and the Information System for Graphological Identification (ISGI)~\cite{mamatsis2023novel} extracts average character shapes using contour and curve detection, after standardizing orientation and size. Such approaches have been leveraged for script or hand classification. 
While they offer visually interpretable outputs well aligned with paleographic reasoning, they typically lack full automation, require extensive preprocessing or manual segmentation, and do not preserve fine-grained spatial information at the level of individual character instances.

More recently, the Learnable Handwriter~\cite{lhw2024} has adapted the Learnable Typewriter~\cite{ltw2023} to learn character prototypes from line images and transcription without requiring any pre- or post-processing. 
We build on this approach to enable automatic and consistent measurements of letter shapes and distances.

\subsubsection{Metrology.}
Early attempts at quantifying graphical features relied on manual measurements, often performed directly on manuscripts or digitized images~\cite{Ornato1997HistoireLivreQuant,muzerelle_gurrado_paleo_systematique,poulle1974paleographie}, as with Léon Gilissen’s `modular ratio' (proportion of letter height and width) to characterize scripts~\cite{gilissen1975rapportmodulaire}. While influential, such approaches suffer from methodological inconsistencies and limited reproducibility~\cite{ornato1975peut-on}. 
Digital tools later enabled semi-automatic measurement of density and letter proportions, such as the Graphoskop~\cite{gurrado2009graphoskop}, which supports the statistical analysis of intra-character and inter-word distances~\cite{gurrado_mesure_2013}. Related statistical approaches have explored the distribution of letter dimensions and spacing to support paleographic comparison~\cite{mcgillivray2005statistical,bischoff1996rythme}. Despite their contribution to objectivity, these methods still depend heavily on manual intervention and predefined measurement protocols.

A further step toward automation was achieved through text–image alignment methods, initially developed for handwritten text recognition. Works such as~\cite{fischer2011transcriptionalignment,bluche2016automatic} use forced alignment techniques -- often based on Hidden Markov Models and Gaussian Mixture Models -- to associate visual features with textual transcriptions and segment text lines vertically into characters. These approaches have been adapted for paleographic tasks, notably to measure inter-letter spacing and average character width~\cite{stutzmann2015text-image-alignment}. However, their primary goal remains character segmentation, with reported error rates on the order of 10–13\%, and they do not naturally lead to consistent instance-level measurements. 

Spatial organization of script, such as inter-character and word spacing, is a neglected field of study in digital paleography. Compact scripts (e.g., `aerated', \textit{scriptio fere continua}) reflect variable and non-standardized word separation, influenced by semiotics, grammar, and character morphology (esp. positional allographs and abbreviations)~\cite{stutzmann2020words}. Studying the arrangement of these elements provides an objective insight into pauses, phrasing, and the overall \textit{rhythm}~\cite{stutzmann2020words} and \textit{density}~\cite{bischoff1996rythme} of a script—features that often escape visual perception.

Our contribution can be seen as bridging morphological and metrological approaches, enabling to automatically obtain consistent measures based on accurate text line alignment with deformed prototypical character shapes.

\begin{figure}[t]
    \centering
    \includegraphics[width=\textwidth]{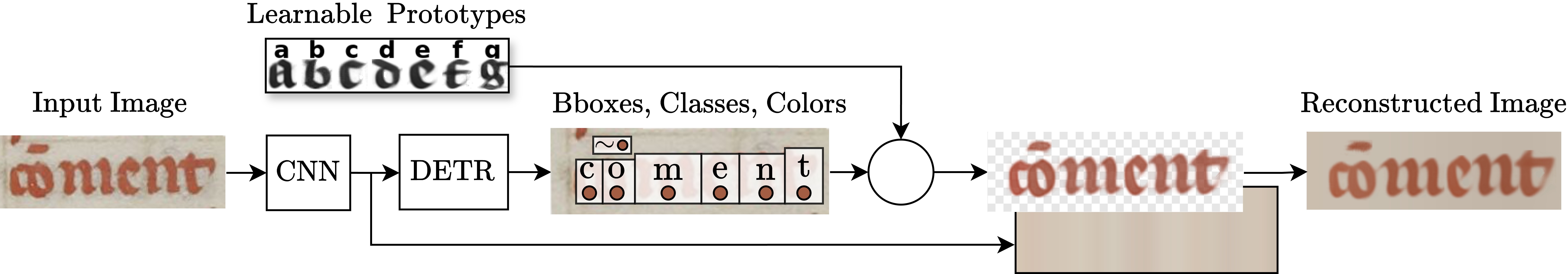}
\caption{\textbf{Overview of the proposed approach}. Building on DTLR, our backbone predicts bounding boxes, class probabilities, and color vectors from text line images. These outputs, alongside learnable prototypes, are used to render characters and compose them spatially onto a predicted background. We leverage bounding box dimensions to accurately affine-render each character.}
    \label{fig:architecture}
\end{figure}

\section{Methodology}
In this section, we first present our detection-based approach to prototypical text line modeling. We then explain how it can be used for paleographic measurements. Finally, we detail the measures and visualizations we use for our case study.

\subsection{Detection-based approach to prototypical text line modeling}

\subsubsection{Architecture.} Our approach is visualized in~\ref{fig:architecture}. Our key idea is to reconstruct text lines from learned prototypical characters, similar to the Learnable-Typewriter~\cite{ltw2023}, but with a detection-based text recognition model, DTLR~\cite{DTLR}, using predicted bounding boxes to deform prototypes. 

We modify DTLR~\cite{DTLR} outputs for each decoder token in two ways. First, to model `accents' (here incl. diacritics and combining characters) and abbreviations, we add to each output token a bounding box regressor and classifiers dedicated to accents and abbreviations, which 
limits the alphabet size. Second, we add linear heads that predict RGB color for each character.

We compute a text line reconstruction in a way similar to the Learnable-Typewriter~\cite{ltw2023}, but taking advantage of our detection backbone. In addition to classification and bounding box prediction, we learn a set of character prototypes defined by grayscale images 
and a small Convolutional Neural Network (CNN) that predicts a low-resolution background image. Our reconstruction has three main differences with~\cite{ltw2023}. First, our detection-based approach prevents multiple use of prototypes to reconstruct a given character, such as an <m> reconstructed using two shapes similar to an <n> reported in~\cite{lhw2024}. Second, we leverage the predicted bounding boxes to change character aspect ratios, while~\cite{ltw2023} was only predicting scale and position.
Third, we take advantage of the character prior learned by the detection approach to modify the image composition procedure to reduce computation time and memory footprint. More precisely, we only composite nearby prototypes in a layered way, and then simply sum masks and colors, which can be efficiently vectorized. This enables a $2.5\times$ reduction of memory footprint. 

\subsubsection{Training Strategy.} We combine the training strategies of DTLR and the Learnable Handwriter~\cite{lhw2024}, leading to a three step strategy. We assume we have a dataset of text lines with their transcription, and that this dataset is split into meaningful subsets that we want to compare, which we refer to as {\it unit of analysis}. 
First, we pretrain our architecture on synthetic data using standard DINO-DETR~\cite{DINO-DETR} detection losses to help the network learn accurate character bounding boxes. We use synthetic data similar to DTLR~\cite{DTLR} but with small modifications to enable learning accents. Second, we train a base model on all text lines of the dataset using a combination of DTLR~\cite{DTLR} {modified Connectionist Temporal Classification (CTC)~\cite{CTC} loss} and $L_1$ line image reconstruction loss. 
Third, following the Learnable Handwriter~\cite{lhw2024}, we fine-tune this model with the same loss, freezing all network components but the prototypes and probability prediction heads, on the text lines of each unit of analysis. Note that this last step is necessary for morphological analysis, but not strictly necessary for our measurements, even if it improves character error rate. 


\subsubsection{Implementation and training details.}
The architecture is built on DINO-DETR~\cite{DTLR} with the default Resnet50 backbone. 
The background is predicted from the Resnet50 feature maps by a 2-layer Convolutional Neural Network (CNN) including: (1) a $3\times3$ convolution with 256 channels followed by BatchNorm and a ReLU activation; (2) a $1\times1$ convolution with 3 channels, sigmoid activation and max-pooling over the height. The resulting vector is bilinearly upsampled to the full image resolution and used as background. Character prototypes are learnable grayscale images with a resolution of $48\times 48$. The character set includes 78 characters, 66 letters and 12 additional unicode characters. 

We perform synthetic pre-training for 1 million iterations, using a learning rate of $10^{-4}$, a batch size of 4, an alphabet of 167 symbols, and 3k randomly sampled fonts. During the last two phases, we use a weight of 3 for the reconstruction loss. To train the base model on the full dataset, we first train the prototypes and classifiers for 80k using a batch size of 16, and a learning rate of $10^{-4}$, then the full network for 300k iterations using a batch size of 32, a learning rate of $10^{-4}$ for the prototypes, and a learning rate of $10^{-5}$ for the text recognition architecture. Finally, the last training stage on each unit of analysis is performed during 4k iterations with a batch size of 4, using a learning rate of $10^{-2}$ for the prototypes and $10^{-5}$ for the background and the characteres classifier, while all other components, including color prediction, remain frozen.

\subsection{From bounding boxes to paleographic measurements.}

\subsubsection{Ambiguity of graphic borders.}
Defining the precise borders of a medieval character 
is challenging. 
Many elements, such as underhanging parts (e.g. ‹g›), overhanging loops (e.g. looped ‹l›), extended feet, accessory strokes, fusions of opposite bows (e.g. ‹bc›), and ligatures, further complicate the assessment. As a consequence, any manual or visual determination of character boundaries introduces subjectivity, which directly affects measurements of character dimensions and inter-character or inter-word distances. To reduce such variability, P.~Saenger~\cite{saenger1997space} proposed measuring inter-word distance only between words that begin and end with vertical strokes to ensure consistent borders for comparability. Instead, we propose to define character position and border using the optimal alignment of a specific character instance with its associated prototype, interpreted as the average character. 

\subsubsection{Bounding boxes of average characters as standardized borders.}
A critical aspect of the bounding boxes our algorithm produces is that they are not defined by human annotations, but instead are defined as the ones producing the best alignment between a specific instance of a character and the prototypical character learned over the full dataset of interest. Indeed, 
the bounding box prediction head and the backbone network are frozen during the last step of training, so that the bounding boxes correspond to the optimal alignment with the average character over the full dataset, and are thus defined consistently across different subsets. Note that, opposite to the one of the Learnable Typewriter~\cite{ltw2023}, our alignment via the bounding box enables aspect ratio changes.

In practice, such a definition implies that overextended attack or exit strokes are excluded from the bounding box. This reduction in variability caused by decorative or contextual elongations ensures that measurements capture the stable structure of the character. Such exclusions do not distort paleographic interpretation. As noted by Stutzmann~\cite{stutzmann2020words}, spatial perception in medieval scripts is not determined by the outermost contours of strokes. Thus, we argue that our bounding boxes, defined by optimal alignment, provide an objective, reproducible, and meaningful measure of character boundaries. Moreover, it enables consistent measurement of dimensions and spacing across manuscripts.


\subsubsection{Refined boxes for improved consistency with human perception.} Our learned character prototypes typically fit well their $48\times48$ canvas (see Figure~\ref{fig:comp_protos}). However, the pixels near the border of the canvas might be very light gray, corresponding to parts of the character that are inconsistent between instances. To better match human perception of the character borders, we define a tighter bounding box in this canvas by computing the filtering mask defined in~\cite{lhw2024}. In our experiments, this tight bounding box is on average 2.8 pixels smaller than the canvas. We transport it to specific character instances, using the transformations defined by the bounding boxes provided by our network, and use it instead of our predicted bounding boxes to define measurements. Note that this step is not necessary, does not change our qualitative results, and is simply added to better fit human intuition.

\subsubsection{Discarding errors.}
While the error rate of our network after the last fine-tuning stage is very low, it is still not zero, and it is important to discard errors before computing our metrics, as detailed in the next section. To do so, we consider both the ground truth and predicted transcription of each text line, and align them via dynamic programming to minimize the Levenshtein distance, similar to what is commonly done to compute the Character Error Rate (CER). This enables us to detect errors (insertion, deletion, and replacement), and to discard any measurement neighboring an error, as well as boxes that touch the image border. To further avoid outlier bounding boxes, we exclude, for each unit of analysis and for each character or bigram, bounding boxes whose height or width deviates by more than 4 standard deviations from the unit of analysis mean. This only excludes a very small proportion of the data, 0.1\% for the letters we analyze in our experiments.

\subsection{Paleographic measures and visualizations}

Using the character bounding boxes, we define measurements 
for characters, bigrams, and word separation, and visualize them using two types of plots.

\begin{figure}[t]
    \centering
        \includegraphics[width=\textwidth]{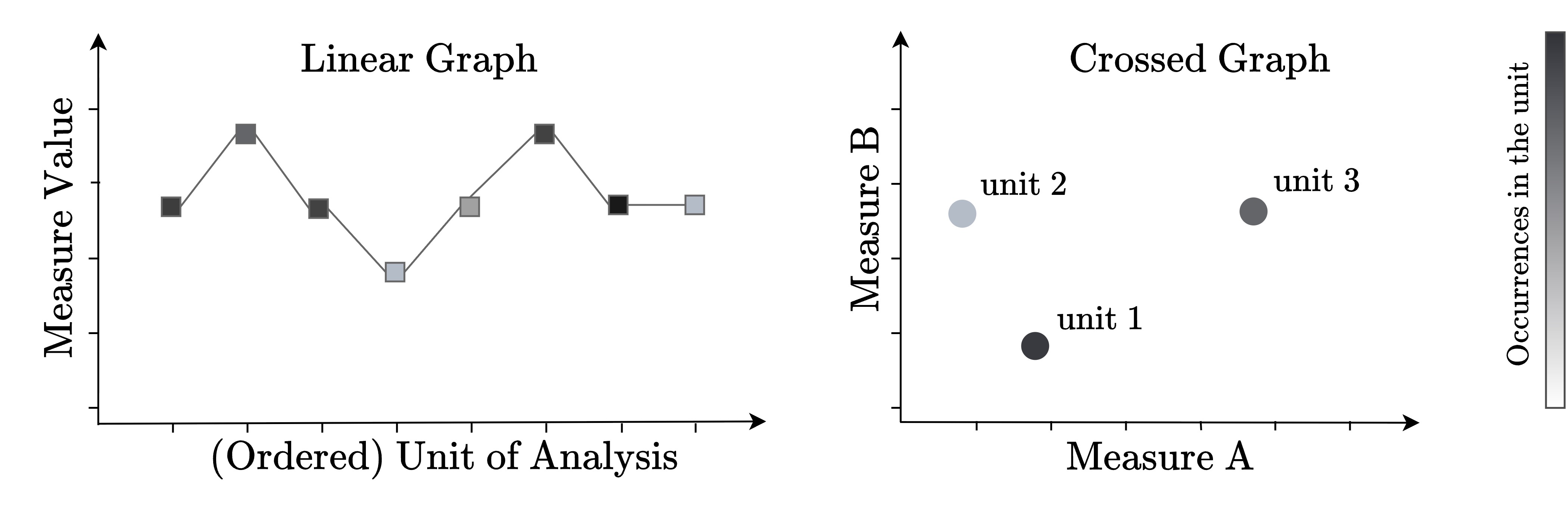}
        \label{fig:dummy_graphs}
        \vspace{-2em}
    \caption{\textbf{Visualization types.}}
    \label{fig:graph_models}
\end{figure}
\subsubsection{Measures.}

To characterize units of analysis, we compute several measures visualized in Figure~\ref{fig:metrics}: width ($\mathbf{w}$) and aspect ratio ($\nicefrac{\mathbf{w}}{\mathbf{h}}$) of specific characters, signed distance between the two characters of a bigram ($\mathbf{d}_b$), aspect ratio of the enclosing bounding box of the bigram ($\nicefrac{\mathbf{w}_b}{\mathbf{h}_b}$) and distance between the last and the first character of consecutive words ($\mathbf{d}_w$). 
For each of these measures, we consider the mean $\mu$ and the {coefficient of variation} $\mathrm{CV} = \frac{\sigma}{\mu}$, where $\sigma$ is the standard deviation, across all instances in the unit of analysis. These measures enable analyzing both character proportionality and horizontal spacing.

In order to obtain an intuitive unit and to compare across manuscripts with different text sizes and image resolutions, we normalize distances using a unit that is invariant to these variables. Inspired by P.~Saenger~\cite{saenger1997space} visual \emph{unit of space}, defined as the distance between minims, we use as reference unit of space half the average width of the lowercase \emph{m} ($\mathbf{w}_{m/2}$).

\subsubsection{Visualisation types.}

We use two types of visualization for these measures, which we refer to as \textit{linear graphs} and \textit{crossed graphs}, illustrated in Figure~\ref{fig:graph_models}. 

Linear graphs display the evolution of a single measure along the units of analysis (e.g., pages, manuscripts ordered by date). These graphs enable us to easily identify trends as well as complementary measures. 

Crossed graphs plot two measures against each other, enabling the analysis of correlations between measures and/or to better differentiate scripts using combinations of complementary measures. 
In both graphs, the intensity of the marker corresponds to the number of occurrences considered in the metric, so as to visually identify less robust data points. 

\section{Experiments}

\begin{figure}[t!]
\centering

\setlength{\intextsep}{6pt}
\setlength{\textfloatsep}{6pt}

\small
\setlength{\tabcolsep}{0pt}
\renewcommand{\arraystretch}{1}

\begin{tabular}{>{\raggedright\arraybackslash}p{0.2\textwidth}
                *{4}{>{\centering\arraybackslash}p{0.2\textwidth}}}
 & \textbf{GP1} & \textbf{GP2} & \textbf{GP3} & \textbf{GP4} \\
\midrule
{Page Range}   & 1r--376r & 265v--457v & 353vb--484ra & 484rb--491v \\
{Units of analysis}         & 65 & 54 & 35 & 6 \\ 
\end{tabular}
\makebox[\textwidth][r]{%
\includegraphics[width=0.8\textwidth]{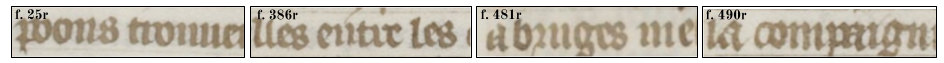}
}
\caption{\textbf{Dataset Summary}, with 160 units of analysis from Paris, BnF, fr.~2813.
}
\label{tab:dataset_gp}

\end{figure}

In this section, we apply our approach to the codex Paris, BnF, fr.~2813. We first introduce the case study and associated data, then discuss the advantages of our model over~\cite{lhw2024} for morphological analysis, and finally present our metrological analysis.

\subsection{Case study and dataset}

We demonstrate our method on the late fourteenth-century {codex} Paris, BnF, fr.~2813— King Charles~V’s personal copy of the \textit{Grandes Chroniques de France}, copied between 1375 and 1379 by four hands~\cite{Hedeman1984ValoisLegitimacy,Oeser1996RaouletOrleans,Peretto2014RaouletOrleans}. It is a luxury manuscript, with a consistent layout of two columns, 50 lines per column, written in a fully separated formal script. Recently,~\cite{vlachou2025grandes-chroniques-fr-2813} used the Learnable Handwriter~\cite{lhw2024} on 68 annotated pages to perform morphological analysis of the four graphic profiles (GP1--GP4) present in the manuscript, complemented by statistical examination of abbreviation usage. This makes this document a suitable case study for demonstrating the benefits of our approach.

We extend the existing annotations~\cite{vlachou_efstathiou_2025grandes-chroniques-fr-2813_dataset} with 92 additional ones, selecting folios to cover successive pages within the same gatherings, the core material unit of production. To avoid distortions introduced by the gutter fold in digitizations, we only consider the outer columns of each page in our analysis. We also omit rubricated lines, which may have been added at a different stage, and truncated lines, for which layout constraints can influence spacing practices. 
This results in a corpus of 160 units of analysis summarized in Figure~\ref{tab:dataset_gp}, totaling \~6,800 transcribed lines and ~280k characters. The dataset is available on Zenodo (DOI: \url{https://zenodo.org/records/18745702}).



\subsection{Comparison to the Learnable Handwriter~\cite{lhw2024}} 

\subsubsection{Performance.}

First, our architecture demonstrates significantly higher memory efficiency than~\cite{lhw2024}. With a batch size of 128, our model requires only 12.7 GB of memory, against 31.3 GB for~\cite{lhw2024}. 

Training is also significantly faster for our method, and we reach convergence -- neither the Connectionist Temporal Classification (CTC)~\cite{CTC} nor the reconstruction loss decreases for 50k iterations-- in 1 day and 20 hours, against 5 days and 2 hours for~\cite{lhw2024} in comparable settings. 
While~\cite{lhw2024} reaches a slightly lower final CER (1.2\% vs. 1.4\% for our method), this better performance comes at the cost of degraded prototypes, which shrink and accumulate contextual noise. This prevents using the converged network of~\cite{lhw2024} for morphological analysis, and requires using an earlier checkpoint associated with a much higher CER of 4.3\%. 

\begin{figure}[t!]
    \centering
    \includegraphics[width=\textwidth]{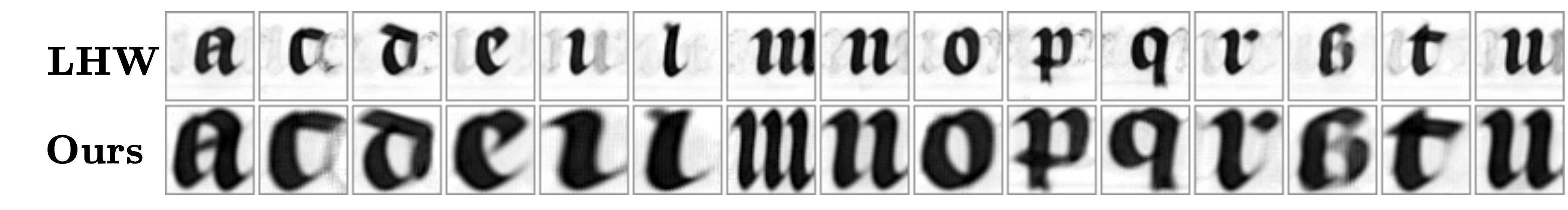}
    \caption{\textbf{Extracted prototypes for Learnable Handwriter (LHW) and Ours} of the 15 most frequent letters for the first page of the dataset (f. 1r).}
    \label{fig:comp_protos}
\end{figure}
\subsubsection{Prototype Quality.}

Figure~\ref{fig:comp_protos} compares the extracted prototypes from our approach with those of~\cite{lhw2024} for the 15 most frequent letters.
As noted in the original paper, even with early stopping,~\cite{lhw2024} incorporates adjacent letter strokes into the prototypes, which must be filtered out for meaningful analysis.
In addition, letters are smaller than the canvas and inconsistently centered. 
By contrast, due to bounding boxes explicitly isolating each character’s body, our prototypes are tightly fitted and well-centered, with fewer artifacts from neighboring characters. 


This can be quantified by using the same strategy to define tight bounding boxes from the base prototypes on both letter prototype sets. The filtering removes an average of 2.8 pixels from our bounding boxes versus 18.4
pixels for~\cite{lhw2024}.

\begin{figure}[t!]
    \centering
    \begin{subfigure}{\textwidth}
        \centering
        \includegraphics[width=\textwidth]{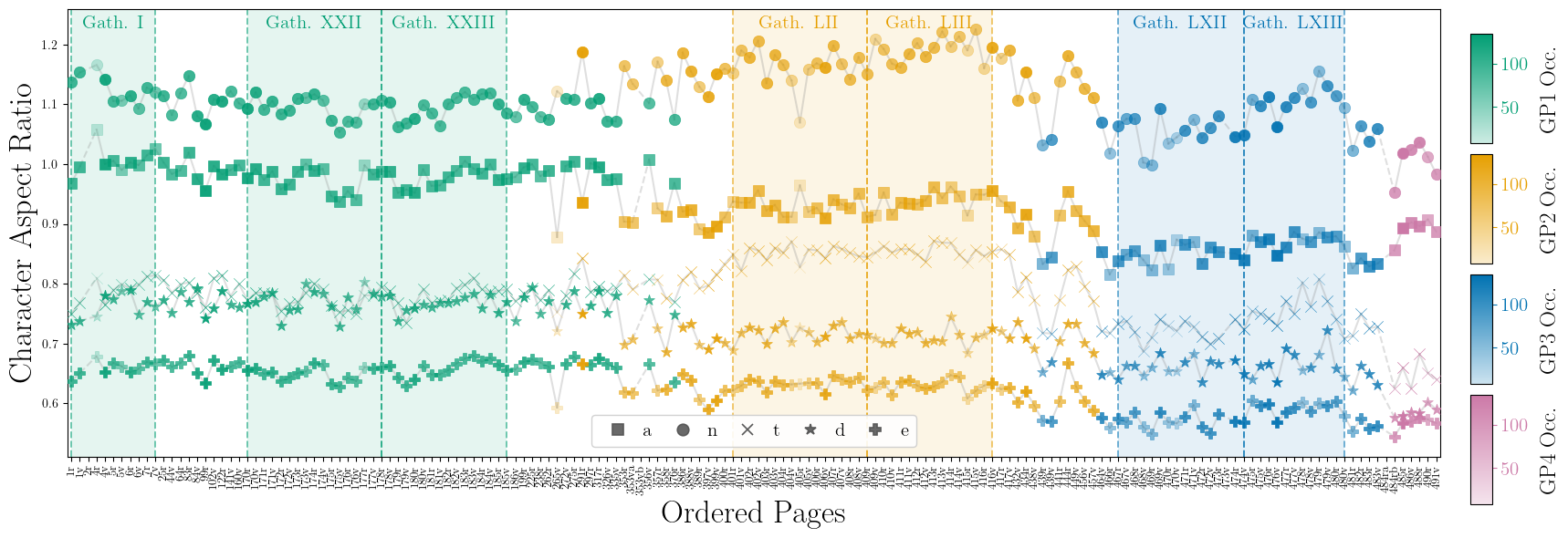}
        \vspace{-1.9em}
        \caption{Characters aspect ratio evolution across the dataset.}
        \vspace{0.5em}
        \label{fig:foliosxmeanAR}
    \end{subfigure}
    
    \vspace{0em}
    
    \begin{subfigure}{\textwidth}
        \centering
        \includegraphics[width=\textwidth]{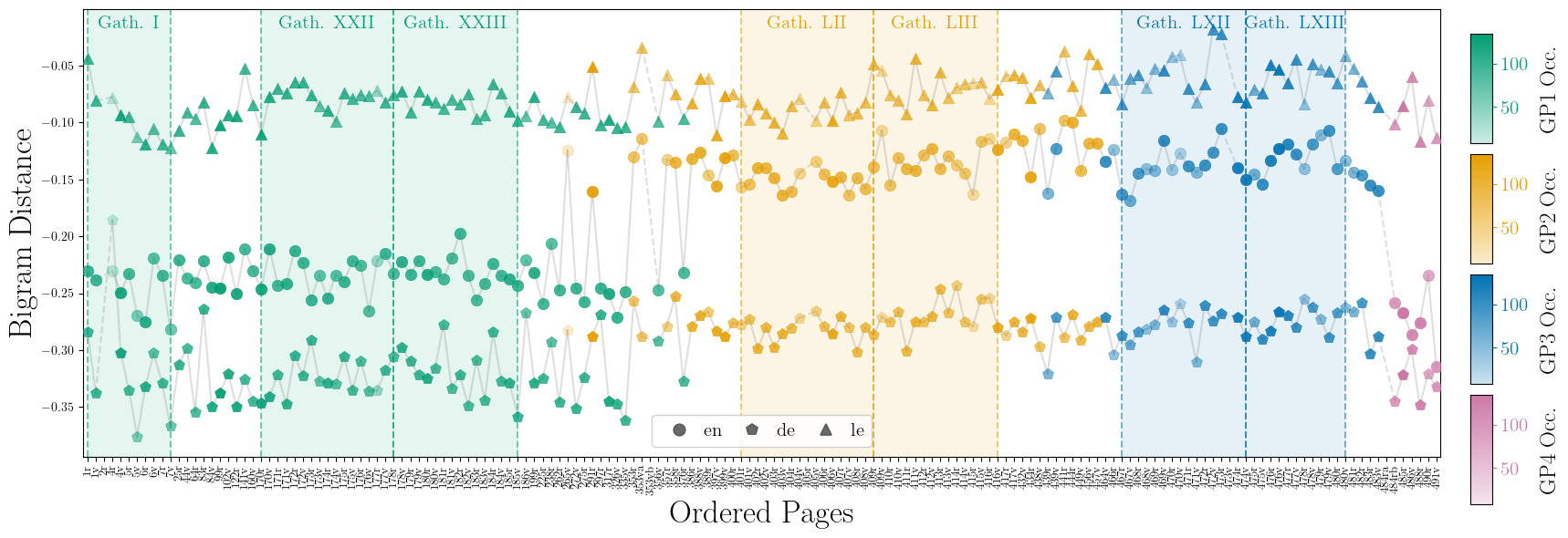}
        \vspace{-1.9em}
        \caption{Bigrams distance evolution across the dataset.}
        \vspace{0.5em}
        \label{fig:foliosxBigram}
    \end{subfigure}
    
    \vspace{0em}
    
    \begin{subfigure}{\textwidth}
        \centering
        \includegraphics[width=\textwidth]{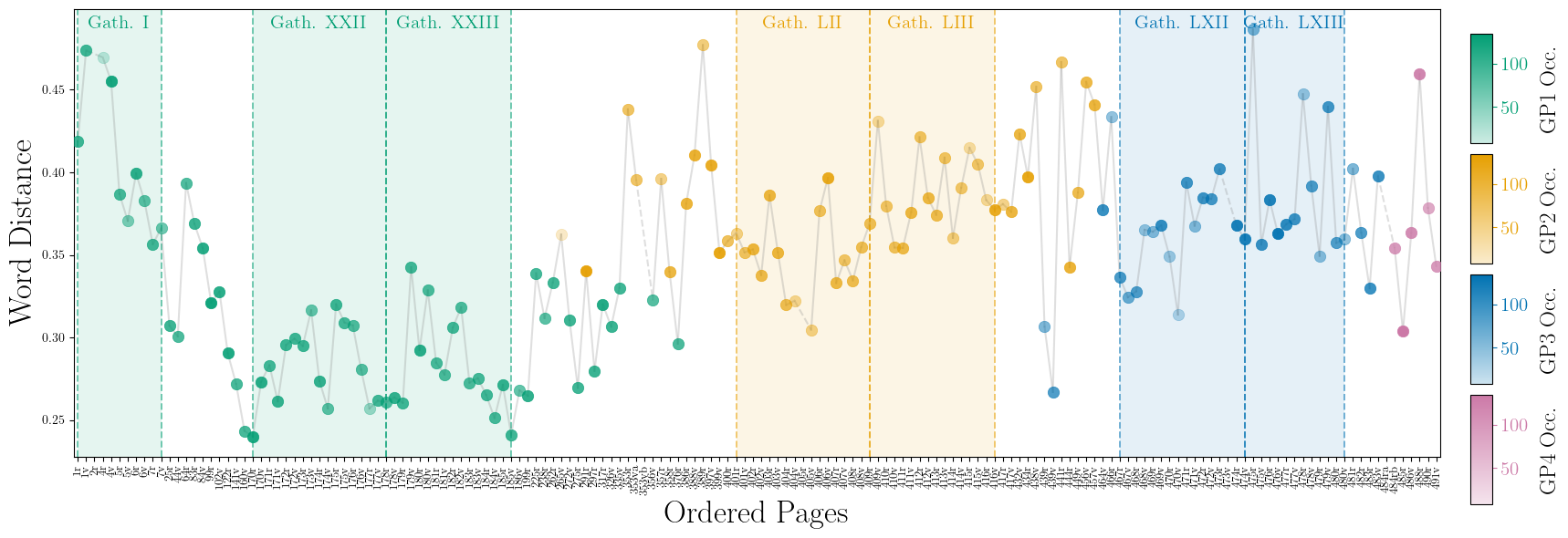}
        \vspace{-1.9em}
        \caption{Word distance evolution across the dataset.}
        \label{fig:foliosxWords}
    \end{subfigure}
    
    \caption{\textbf{Linear Graphs.}  Highlighted regions show consecutive pages. Dotted lines indicate the beginnings and endings of gatherings. We report mean quantities over units of analysis (pages).}
    \vspace{-0.5em}
    \label{fig:folios_combined}
\end{figure}

\begin{figure}[t!]
    \centering
    

    \begin{subfigure}[b]{0.48\textwidth}
        \centering
        \includegraphics[width=\textwidth]{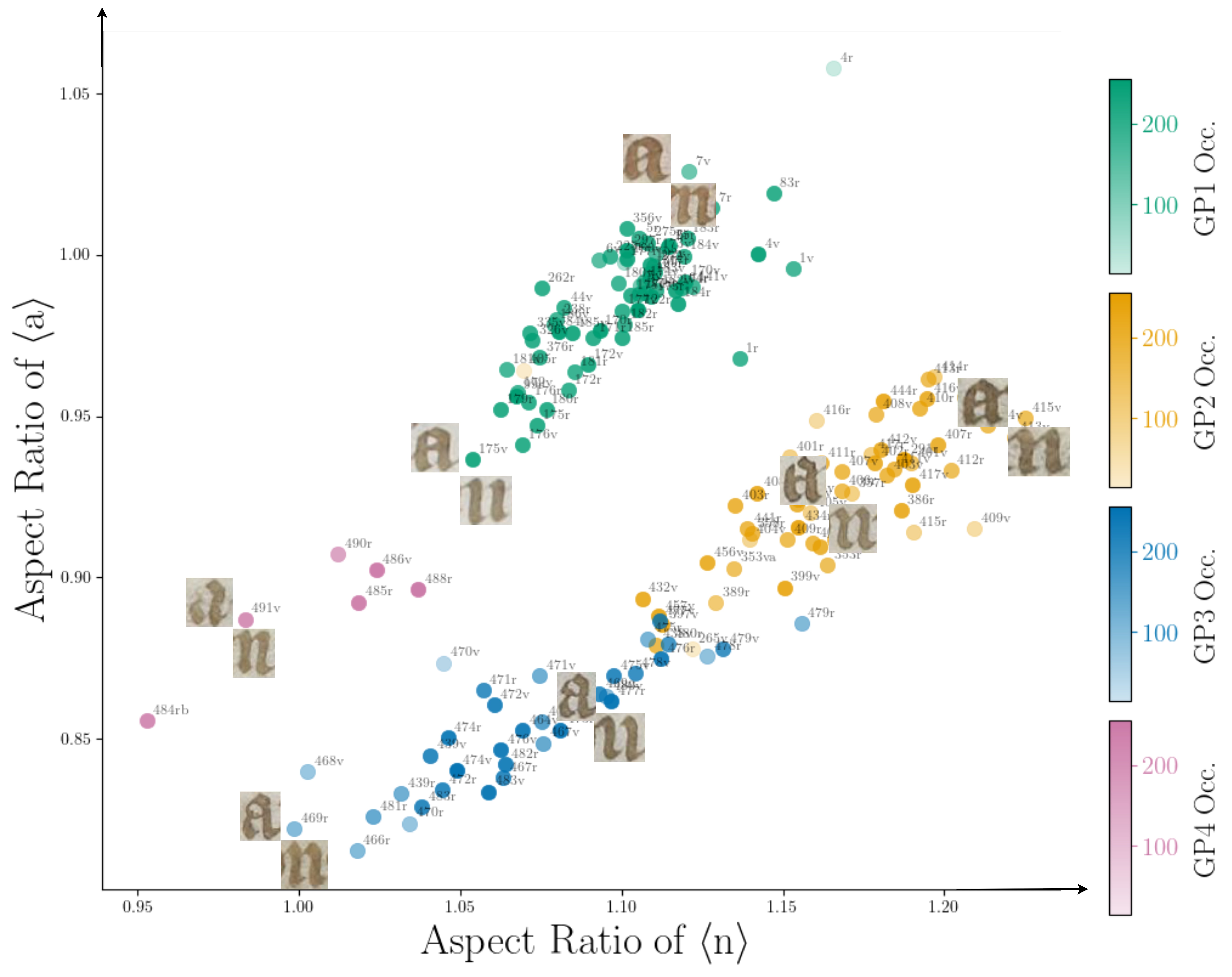}
        \caption{$\mu$[AR] of ‹a› × $\mu$[AR] of ‹n› }
        \label{fig:meanARaxmeanARn}
    \end{subfigure}
    \hfill
    \begin{subfigure}[b]{0.48\textwidth}
        \centering
        \includegraphics[width=\textwidth]{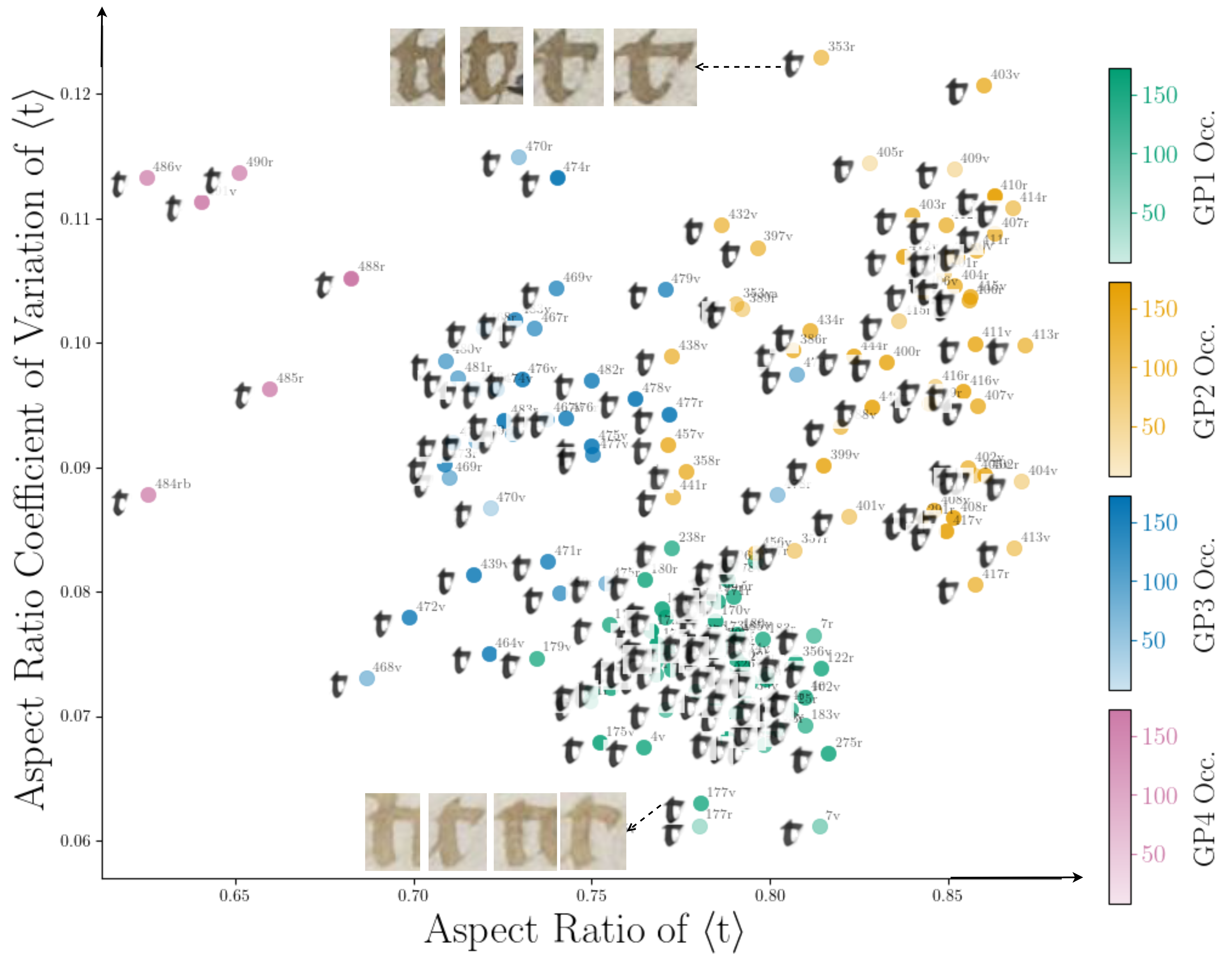}
        \caption{letter ‹t›: CV[AR] × $\mu$[AR] }
        \label{fig:meanARtaxCV}
    \end{subfigure}
    \vspace{0em}
    
    \begin{subfigure}[b]{0.48\textwidth}
        \centering
        \includegraphics[width=\textwidth]{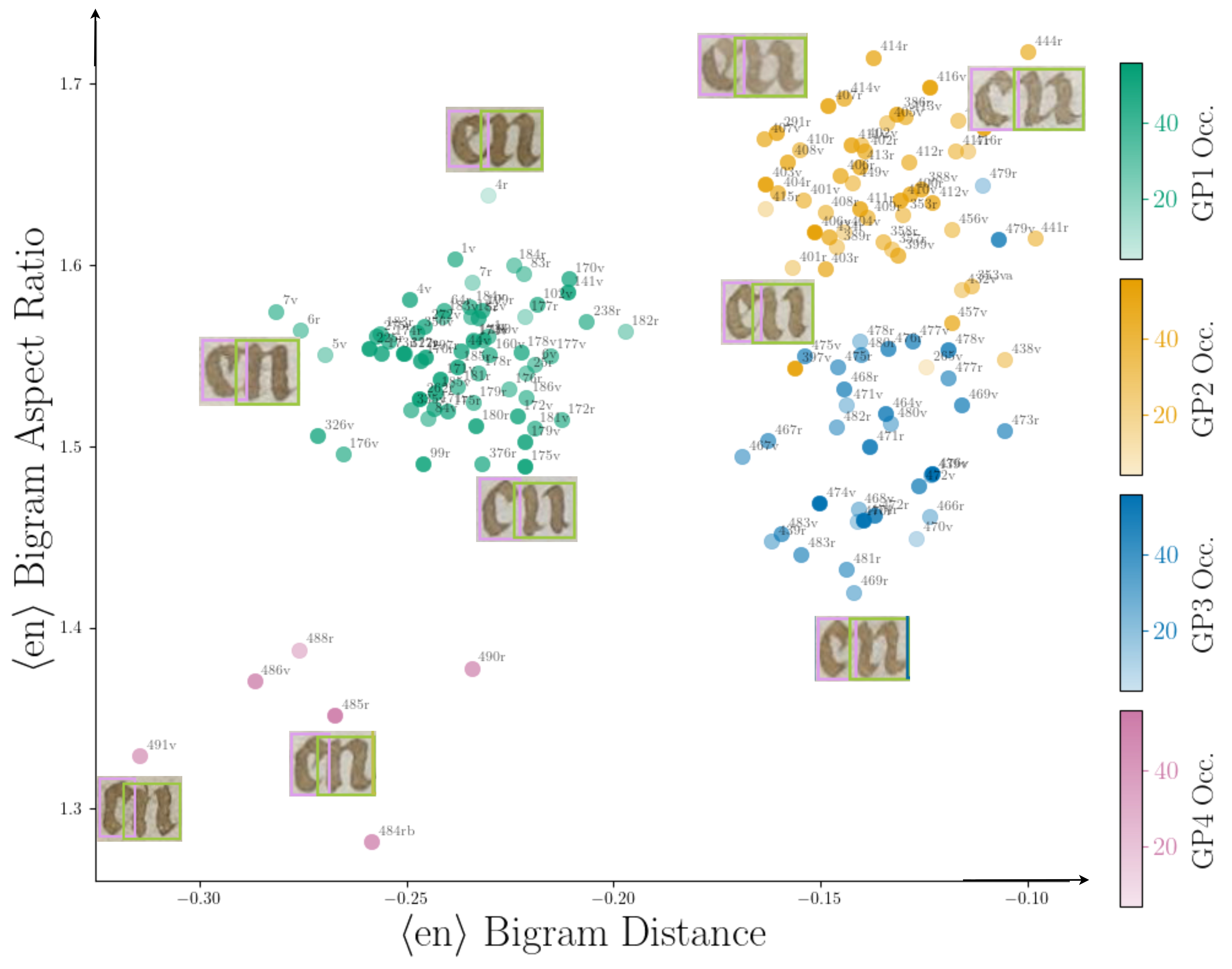}
        \caption{Bigram ‹en›: $\mu$[$d_b$] × $\mu$[AR]}
        \label{fig:bigramenxNormDistxAR}
    \end{subfigure}
     \begin{subfigure}[b]{0.48\textwidth}
         \centering
        \includegraphics[width=\textwidth]{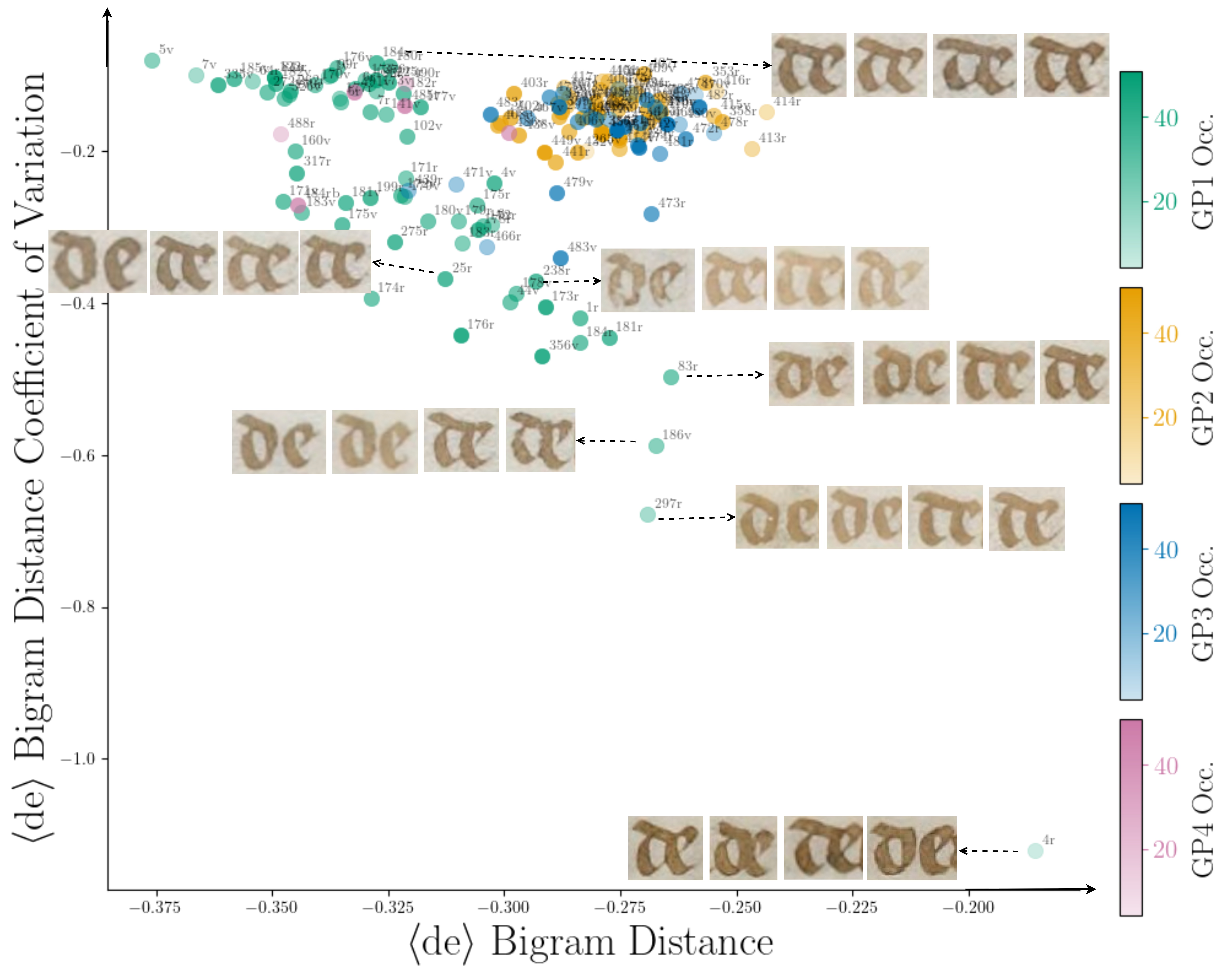}
        \caption{Bigram ‹de›: CV[$d_b$] × $\mu$[$d_b$] }
        \label{fig:bigramdexNormDistxCV}
    \end{subfigure}
    
    \caption{\textbf{Crossed graphs for characters and bigrams.} We cross information between the mean ($\mu$) and coefficient of variation (CV) of different letters and bigrams aspect ratio (AR) and distances between bigrams characters ($d_b$). }
    \label{fig:AR_subfigures}
\end{figure}

\subsection{Novel metrological analysis}
In this section, we analyze our results in two complementary dimensions of script characterization: proportionality and horizontal compression. 
Figure~\ref{fig:folios_combined} shows linear graphs for selected letters' aspect ratios, selected bigrams' distances, and spaces between graphical units. Figure~\ref{fig:AR_subfigures} shows crossed graphs for different measures discussed below. The advantage of our approach is, of course, that such graphs can easily be produced across many units of analysis and for many characters and measures once our models have been trained. 

\subsubsection{Letters proportionality and consistency.} 
Figure~\ref{fig:foliosxmeanAR} provides an overview of the evolution of letter aspect ratios across the dataset. The four Graphic Profiles (GPs) are already distinguishable on this basis alone, an observation further supported by the pairwise comparison of complementary letters such as ‹a› and ‹n› in Figure~\ref{fig:meanARaxmeanARn}. In this plot, the aspect ratio of ‹a› is plotted against that of ‹n›, revealing structured proportional relationships within each Graphic Profile. Two coherent proportional regimes emerge: one shared by GP1 and GP4, and another shared by GP2 and GP3, each associated with a distinct proportionality factor. This suggests differentiated internal organisation across profiles, potentially reflecting differences in letter construction techniques.

To further examine inter- and intra-profile letter variation, we relate the aspect ratio of ‹t› to its coefficient of variation in Figure \ref{fig:meanARtaxCV}. The coefficient of variation provides a measure of executional consistency within each profile. GP1 exhibits the lowest variability, forming a compact cluster that indicates strong intra-profile stability; GP4 shows a comparable clustering pattern, albeit with a higher coefficient of variation. By contrast, GP2 and GP3 display greater variability. Examination of the manuscript suggests that, for GP2, this dispersion reflects its tendency to elongate the horizontal bar of ‹t› in line-final positions. In comparison, GP1 maintains a stable morphological configuration across contexts, whether ‹t› occurs in ligature, in isolation, or at the ends of lines.

\subsubsection{Horizontal compression of the graphic chain.}
We now analyze the horizontal compression of the graphic chain, both inside bigrams and between words. 

In Figure~\ref{fig:foliosxBigram}, we plot the average inter-character distance on our three selected bigrams, and in Figure~\ref{fig:foliosxWords}, the average inter-word distance. 
The distances between bigrams are consistently negative, indicating systematic overlap of bounding boxes, while the distances between words are positive. While there are differences between GPs, inter-character bigram distances alone do not seem enough to completely differentiate them, GP2 and GP3 showing, for example, very similar characteristics. There are also important variations within some GPs. In particular, the distances between words decrease significantly from the beginning of GP1's copying, stabilizing thereafter. 
The discovery of such a subtle effect, which in this case might be related to an effort towards a more efficient space management strategy after a more lavish beginning, outlines the power of our analysis tool.

Distances can, of course, be crossed with other measures, such as aspect ratio -- as illustrated in Figure~\ref{fig:bigramenxNormDistxAR} for the ‹en› bigram -- to enable graphic profile separation. Unlike our observation for letters, there is no clear proportionality between this specific bigram aspect ratio and its (inter-letter) distance, suggesting that spacing practices do not map directly onto letter morphology.

The distance in the ‹de› bigram is the smallest in all GPs, which can be related to the fusion of opposite bows, known as the first rule of Meyer~\cite{Zamponi1988Elisione}. 
In Figure~\ref{fig:bigramdexNormDistxCV}, we further analyze this example by crossing the bigram distance with its coefficient of variation. Note that the coefficient of variation is negative because the mean distance is negative, so the most negative values correspond to the greater variation within a unit of analysis. Interestingly, the coefficient of variation is similar for GP2, GP3, GP4, and part of GP1, but varies much more in GP1, sometimes taking very negative values, indicating important variance, with only a small correlation with larger average distances. At first view, this is surprising because GP1 constitutes the most consistent graphic profile. A closer qualitative examination enables, however, to understand this effect: while ‹de› is predominantly executed with the opposite bows of ‹d› and ‹e› fused, in specific pages, it appears with an intervening space, mostly as a preposition or at the beginning of a word (e.g., ‹des-›). This alternative spacing occurs almost exclusively in line-final positions, a pattern not observed in the other GPs. Again, this discovery of a relatively rare and inconsistent effect is far from trivial, outlining the power of our systematic measurements and analysis graphs. 

\section{Conclusion}

In this work, we presented a framework for paleographical analysis that unifies morphological and metrological approaches to enable scalable measurements. We rely on an architecture and training strategy that bring together detection-based text recognition and prototype-based image reconstruction. We show how explicit character detection leads to more stable and efficient training, higher-quality character prototypes and scalable meaningful paleographic measurements. 

Our case study on the \textit{Grandes Chroniques de France} (codex Paris, BnF, fr.~2813) demonstrates how our measures and visualizations enable an in-depth study of proportionality and horizontal compression. Indeed, beyond providing many criteria for hand separation, our visualizations led to the discovery and analysis of subtle paleographic phenomena. 

\section*{Acknowledgment}
This study was supported by the CNRS through MITI and the 80|Prime program (CrEMe), and by the European Research Council (ERC project DISCOVER, number 101076028). This work was granted access to the HPC resources of IDRIS under allocation AD010614956R1 and AD011015222, made by GENCI.

%
%
%
\bibliographystyle{splncs04}
\bibliography{bibliography}

\end{document}